\documentclass[journal]{elsarticle}
\usepackage[utf8]{inputenc}
\usepackage[flushleft]{threeparttable}
\usepackage[utf8]{inputenc} 
\usepackage[T1]{fontenc}    
\usepackage{hyperref}       
\usepackage{url}            
\usepackage{booktabs}       
\usepackage{amsfonts}       
\usepackage{nicefrac}       
\usepackage{microtype}      
\usepackage{amsmath}
\usepackage{graphicx}
\usepackage[ruled,vlined]{algorithm2e}
\usepackage{booktabs}
\newcommand{\vcl}{VCL}
\newcommand{\VIPARR}{RCL-CPB}
\usepackage{multirow}
\usepackage{subfigure}
\usepackage[export]{adjustbox}
\usepackage{makecell}
\usepackage{siunitx}
\usepackage{caption}
\begin{document}
\begin{frontmatter}

\title{Robust Continual Learning through a Comprehensively Progressive Bayesian Neural Network}

\author{Yang Guo, Cheryl Sze Yin Wong and Ramasamy Savitha\\
  Insitute for Infocomm Research\\
  Agency for Science, Technology and Research\\
  Singapore \\
  \texttt{\{guo\_yang,cheryl\_wong,ramasamysa\}@i2r.a-star.edu.sg}
}

\begin{abstract}
  This work proposes a comprehensively progressive Bayesian neural network for robust continual learning of a sequence of tasks. A Bayesian neural network is progressively pruned and grown such that there are sufficient network resources to represent a sequence of tasks, while the network does not explode.  It starts with the contention that similar tasks should have the same number of total network resources, to ensure fair representation of all tasks in a continual learning scenario. Thus, as the data for new task streams in, sufficient neurons are added to the network such that the total number of neurons in each layer of the network, including the shared representations with previous tasks and individual task related representation, are equal for all tasks. The weights that are redundant at the end of training each task are also pruned through re-initialization, in order to be efficiently utilized in the subsequent task. Thus, the network grows progressively, but ensures effective utilization of network resources. We refer to our proposed method as 'Robust Continual Learning through a Comprehensively Progressive Bayesian Neural Network (RCL-CPB)' and evaluate the proposed approach on the MNIST data set, under three different continual learning scenarios. Further to this, we evaluate the performance of \VIPARR~on a homogeneous sequence of tasks using split CIFAR100 (20 tasks of 5 classes each), and a heterogeneous sequence of tasks using MNIST, SVHN and CIFAR10 data sets. The demonstrations and the performance results show that the proposed strategies for progressive BNN enable robust continual learning.
\end{abstract}
\begin{keyword}
Continual learning \sep Bayesian Neural Network \sep Progressive Neural Network \sep Evolving Architecture 
\end{keyword}
\end{frontmatter}

\section{Introduction} \label{sec:intro}
Recently, there is tremendous progress in the adoption of deep learning models for real-world applications, where data streams in with an evolving data characteristic. Although traditionally trained deep neural networks are capable of representing complex functional relationships, they have a strong assumption that the testing data belongs to the distribution of the training data, and are not capable of adapting to the needs of streaming data. Continual learning is a subset of lifelong learning approaches that is aimed at learning such streaming data as they are available, without catastrophically forgetting any distribution that it has learnt \cite{liubing2020}. Each new distribution or new class of data is referred to as a task, and such tasks can be characterized with domain incremental (where the distribution of data drifts with increasing tasks) and/or class incremental drifts (where the number of output classes evolves with increasing tasks). Addressing these drifts enables deep neural networks to learn and represent several tasks robustly.

Bayesian inference provides a natural framework for continual learning, as it has the ability to avoid catastrophic forgetting through the use of the posterior distribution of the trained weights, while learning the likelihood of the oncoming data \cite{vcl}. Furthermore, the uncertainty estimates of the Bayesian Neural Network (BNN) helps to provide reliability estimates \cite{uncertainty_transp2020} about the model and its representations, which can be used to establish the significance of individual weight parameters. Therefore, variational inference \cite{vcl} and uncertainty estimates of BNN \cite{ebrahimi2019uncertaintyguided} have been used to regularize network representations towards continual learning. However, robust representation of a long sequence of tasks requires providing sufficient network resources for all tasks, while ensuring that the resources are prudently shared across multiple tasks. Thus, there is a need to make structural changes to the BNN, while regularizing representations based on the variational inference and uncertainty estimates, to improve robustness in continual learning.

In this paper, we propose a comprehensively progressive Bayesian Neural Network (\VIPARR) for robust continual learning. It starts with the contention that similar tasks should have same number of total network resources, to ensure fair representation of all tasks \cite{PNN}. At the end of training a task, the significance of individual weight parameters in the Bayesian neural network are computed and the redundant weight connections are pruned. With the onset of data for a new task, the number of network resources that can be shared with the previous tasks are estimated using the mean activations of the neurons in individual layers of the network for the current task. Thereafter, the number of neurons to be added in each layer of the Bayesian neural network is computed based on these estimations. This is the first time in the literature of continual learning where the shared representations and the required network resources for individual task related representations are estimated in order to decide the number of neurons to be added to the individual layers of the Bayesian neural network. This is so unlike the earlier methods with architectural strategy \cite{PNN}, \cite{DEN}, \cite{NIPS2019_9518} with iterative pruning and/or addition of network resources.

These structural changes, in addition to the regularization of the weights for the Bayesian neural network \cite {vcl} and replay through coreset samples, help to learn a sequence of tasks very robustly. Regularization is aimed at regularizing the network parameters through KL-divergence to ensure the past tasks are not catastrophically forgotten, while learning a new task through minimizing the cross-entropy loss. The replay helps to remind the network of the past tasks, through a saved subset of samples from each task. The proposed approach is therefore referred to as, 'Robust Continual Learning through a Comprehensively Progressive Bayesian Neural Network' (\VIPARR). The proposed \VIPARR~is evaluated against state-of-the-art continual learning methods in domain incremental and class incremental scenarios. Specifically, we use the permuted MNIST, split MNIST, CIFAR100 data sets to demonstrate the ability of \VIPARR~to learn a sequence of homogeneous tasks. Furthermore, its ability to learn heterogeneous tasks is demonstrated using a sequence of MNIST, SVHN, and CIFAR10 data sets. In addition, ablation studies to emphasize the significance of individual strategies are performed. Performance studies show the robustness of the proposed \VIPARR~in learning homogeneous and heterogeneous sequence of tasks.

The main contributions of the paper are summarized below:
\begin{itemize}
    \item We develop a Robust Continual Learning approach through a Comprehensively Progressive Bayesian Neural Network (RCL-CPB) that estimates the shared representations for a task with the existing network resources in a Bayesian neural network. 
    \item For the first time in the literature of continual learning, the proposed \VIPARR~estimates the number of neurons to be added to the individual layers of the BNN, while the data for a new task streams in. This enables non-iterative pruning and addition of neurons, which is an improvement to the traditional progressive neural network.
    \item The performance of the proposed \VIPARR~is studied on a number of data sets with homogeneous sequence of tasks and heterogeneous sequence of tasks.
    \item The advantages of the individual strategies of continual learning are demonstrated using ablation studies.
\end{itemize}

The paper is organized as follows: the related literature in continual learning is summarized briefly in Section \ref{related}. We introduce the preliminaries of Bayes by Backprop and continual learning in Section \ref{prelim}. Section \ref{sec:method} presents the proposed \VIPARR~. We present the results and demonstrations using the experimental study on the various tasks (homogeneous and heterogeneous data sets) in Section \ref{sec:results}. Further, the effect of individual strategies are studied, and the analysis is also presented in Section \ref{sec:results}. Finally, the conclusions of the paper are summarized in Section \ref{sec:conclusion}.

\section{Related Works}\label{related}
Continual learning algorithms have been developed using architectural, regularization and replay strategies in the literature \cite{clreview,delange2021clsurvey}. The architectural strategies impart structural changes to the neural network architecture, by growing, pruning and/or freezing different parts of the network for different tasks. Progressive Neural Networks (PNN) \cite{PNN}\cite{cl_progressivedeep}, Dynamically Expandable Networks \cite{DEN}, Compacting, Picking and Growing \cite{NIPS2019_9518} and Learn to Grow \cite{learntogrow} are a few examples of continual learning algorithms using architectural strategies. However, these algorithms increase the complexity of network and/or computations \cite{PNN} \cite{DEN} \cite{NIPS2019_9518} \cite{learntogrow}. Regularization strategies constrain the weight adaptations of the model for various tasks to reduce catastrophic forgetting. Elastic Weight Consolidation (EWC) \cite{EWC2017}, Learning without Forgetting (LwF) \cite{lwf}, Synaptic Intelligence (SI) \cite{si}, and ensemble methods \cite{ensemble2017}, are some continual learning algorithms using regularization strategies for continual learning. However, despite the increased computations due to additional loss terms to avoid catastrophic forgetting in regularization based approaches, their performance on old and new tasks could be compromised due to a limited amount of neural resources \cite{clreview}. The replay strategies leverage on replaying data from previous tasks to reinforce and remind the model of earlier tasks, through one of the following ways: (1) saving a subset of data from past tasks \cite{GEM,icarl} and/or (2) capturing representations of past tasks for generative \cite{pellegrini2019latent,Shin2017ContinualLW} or constructive replay \cite{flashcards2020} or dual memory frameworks \cite{dual_memory}. The combinations of these three strategies of continual learning tend to be complementary \cite{Parisi2020}, and there are also algorithms to combine various strategies such as AR1 \cite{AR2018}. The AR1 \cite{AR2018}, which is a combination of architectural and regularization strategies, does not adapt well to larger incremental classes. In addition to the above-mentioned issues, all these methods are not transparent and do not provide confidence of representations and inference. 

On the other hand, variational inference based approaches such as Variational Continual Learning (VCL) \cite{vcl} and Uncertainty based Continual Learning (UCB) \cite{ebrahimi2019uncertaintyguided} have also been proposed in the literature. In VCL, regularization between old and new tasks is performed by using the posterior distribution of the old task as the prior distribution of the new task. While in UCB, the significance score of weight parameter is either used to adapt the learning rate for each weight parameter or used to prune insignificant weight. However, both these methods leverage on an {\it a priori} fixed BNN architecture. Thus, when the number of tasks increases and gets more complex, these methods do not provide sufficient network resources for learning continually. Recently, generalized VCL (GVCL) \cite{loo2021generalized} explored the effect of KL divergence in VCL and found that VCL with over-emphasis on KL-divergence results in over-regularization, limiting the network resources for the new task. This highlights that there is a need for increasing network resources in continual learning tasks, particularly in VCL, as emphasized in \cite{bsa_cl}. However, unlike the iterative process of estimating the number of neurons in previous works of progressive continual learning, we propose a comprehensively progressive Bayesian neural network for robust continual learning.

\section{Preliminaries}\label{prelim}
In this section, brief introductions on continual learning and Bayesian neural networks are provided in Sections \ref{sec:cl-intro} and \ref{sec:bnn-intro}, respectively.

\subsection{Continual Learning}\label{sec:cl-intro}
Continual learning is aimed at robust learning of a sequence of tasks, such that the model is able to provide accurate inference on all tasks. Typically, the data becomes unavailable after being trained, hence the model has to learn in an incremental manner. Such increments can occur either as a drift in the distribution of data (domain incremental drifts) or in the number of classes (class incremental drifts) in the data. 

Let us assume that there is a sequence of $l$ tasks ($T_1, \ldots T_l$), and the data for each task is given by $\mathcal{D}_{t};~ t = 1,\ldots, l$. Let the number of samples in each task be $N_t$, such that $\mathcal{D}_{t} \in \Re^{N_t}\times m$, and each sample occurs in input-output pairs $(\mathbf{x}^j_{t},c^j_{t}); j=1,\ldots,N_t$. The input $\mathbf{x}^j_{t} \in \Re^m$ and the class labels $c^j_{t} \in \left\{1,\ldots,nc\right\}$ of each task may be assumed to be represented by $m$ features, and $nc$ classes. It must be noted that not all tasks may have the same $m$, $nc$ and $N_t$. The class labels are one-hot encoded to derive the target outputs $\mathbf{y}^j_{t} \in [0,1]^{nc}$. Thus, the data for all tasks 
$\mathcal{D} = [\mathcal{D}_{1} \ldots \mathcal{D}_{t} \ldots \mathcal{D}_{l}]$ occur sequentially, and the objective of the continual learning technique is to progressively learn this sequence of tasks without forgetting the past tasks.


\subsection{Bayesian Neural Network}\label{sec:bnn-intro}
In this section, we briefly introduce Bayes by Backprop \cite{wtunce2015}, which is used to train the Bayesian neural network in this paper. 

In a BNN, weights are learnt as probability distributions, instead of fixed values. 
Let the weights of the BNN be $\mathbf{w}$ and the data for task $t$ be $\mathcal{D}_{t}(\mathbf{x}_{t}, \mathbf{y}_{t})$. The objective of the BNN is to learn the true posterior distribution of weights $p(\mathbf{w}|\mathcal{D}_{t})$. As $p(\mathbf{w}|\mathcal{D}_{t})$ is intractable, variational inference is used to approximate the distribution $p(\mathbf{w}|\mathcal{D}_{t})$ with $q(\mathbf{w}|\mathbf{\theta})$ parameterized by $\mathbf{\theta}$ and minimizing the Kullback-Leibler (KL) divergence between $q$ and $p$.
\begin{eqnarray}
\theta^{*} & =& \arg\min_{\theta} \text{KL}[q(\mathbf{w}|\mathbf{\theta})||p(\mathbf{w}|\mathcal{D}_{t})]
\label{eq1}
\end{eqnarray}

The loss function can be derived from equation\eqref{eq1} as below:
\begin{equation}
    \mathcal{L}(\mathcal{D}_{t}, \mathbf{\theta}) =  \text{KL}[q(\mathbf{w}|\mathbf{\theta})||p(\mathbf{w})] - \mathbb{E}_{q(\mathbf{w}|\mathbf{\theta})}[\log p(\mathcal{D}_{t}|\mathbf{w})]
\label{equa_loss}
\end{equation}

Equation\eqref{equa_loss} can be approximated through $M$ Monte Carlo sampling as below:
\begin{equation}
\mathcal{L}(\mathcal{D}_{t}, \mathbf{\theta}) \approx \sum_{i = 1}^{M} \log q(\mathbf{w}^{(i)}|\mathbf{\theta}) - \log p(\mathbf{w}^{(i)}) - \log p(\mathcal{D}_{t}|\mathbf{w}^{(i)})
\end{equation}

In the continual learning scenario, the prior distribution p($\mathbf{w}$) of the first task is set to zero-centered Gaussian distribution and prior distributions for subsequent tasks are set based on the posterior distribution from the previous task \cite{vcl}. Following the setting from \cite{vcl} for fair comparison, we assume that $q(\mathbf{w}|\mathbf{\theta})$ also follows a Gaussian distribution parameterized by $\mathbf{\theta}(\mathbf{\mu}, \mathbf{\rho})$ with mean $\mathbf{\mu}$ and variance $\mathbf{\sigma^{2}} = \exp{(\rho)}$. Thus the standard deviation can be represented as $\mathbf{\sigma} = \exp{(\rho\times 0.5)}$ to ensure that $\mathbf{\sigma}$ is always positive. Reparameterization trick is used to sample weight from the variational posterior $\mathbf{w} = \mathbf{\mu} + \mathbf{\sigma} \circ \mathbf{\epsilon}$ where $\circ$ represents element-wise multiplication and $\mathbf{\epsilon}$ is a random sample from $\mathcal{N}(\mathbf{0},\mathbf{I})$. In our experiments, $\mathbf{\mu}$ is initialized with $\mathcal{N}(\mathbf{0},\mathbf{0.1})$ and $\mathbf{\rho}$ is initialized as constant ($-6$ or $-3$ dependent on the experiment).

\section{A Comprehensively Progressive Bayesian Neural Network for Robust Continual Learning} \label{sec:method}
\begin{figure*}
  \centering
  \includegraphics[width=\textwidth]{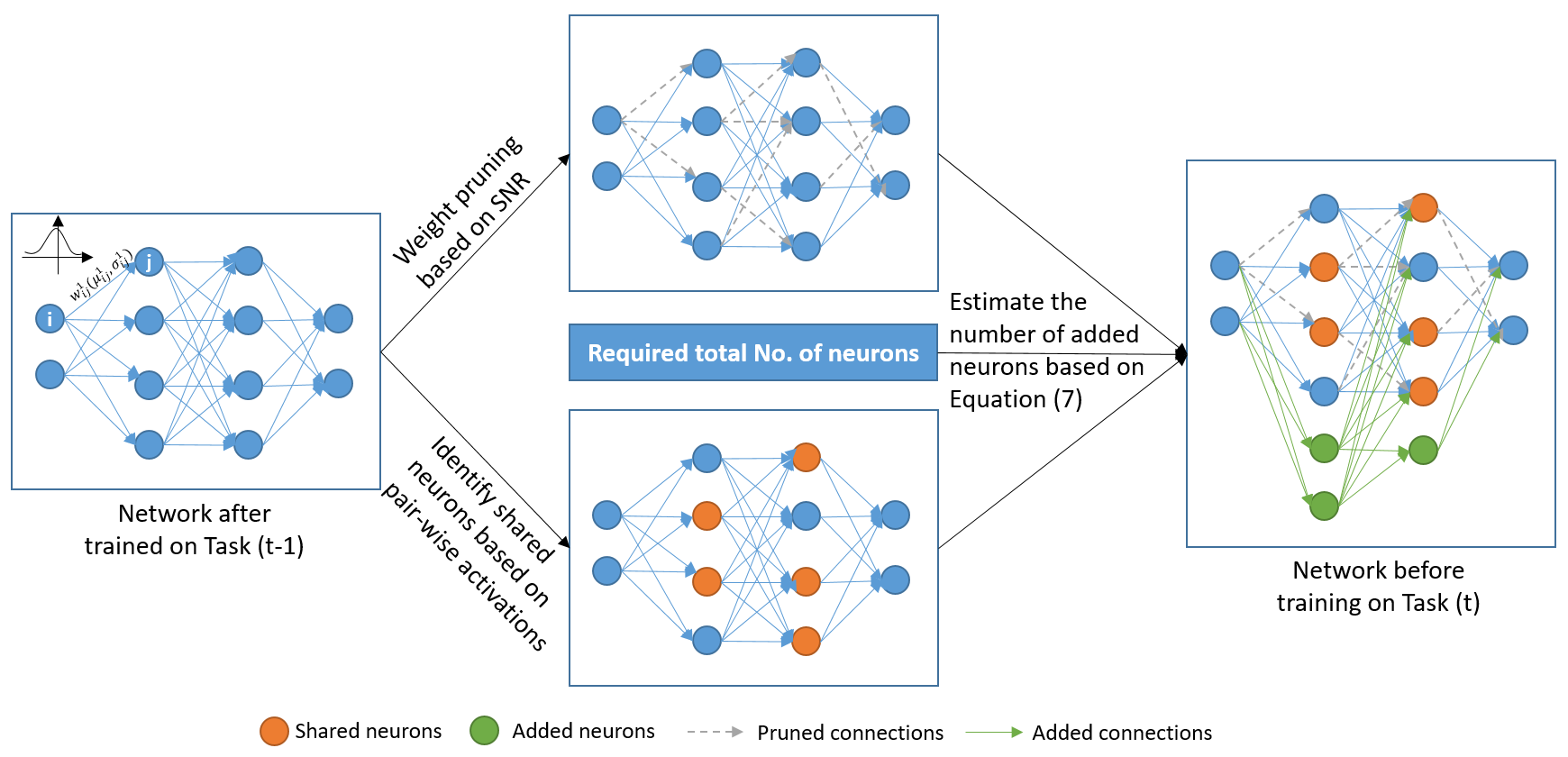}
  \caption{Robust Continual Learning through a Comprehensively Progressive Bayesian Neural Network}
  \label{fig:framework}
\end{figure*}

In this section, we elaborate the robust continual learning approach through a comprehensively Progressive Bayesian neural network, depicted in Fig. \ref{fig:framework}. The \VIPARR~progressively grows the individual layers of a Bayesian neural network on as-needed basis. To this end, it prunes the neurons in individual layers of the BNN at the end of each task, and adds sufficient neurons based on the novelty and resource requirement of the new task, as explained in this section.

Without loss of generality, let us assume that the network has been trained on $\left(t-1\right)$ tasks, and is presented with the data of task $t$, $D_{t}$. Let us assume that the network has $o$ hidden layers, with $K_o$ neurons in each layer. 

The comprehensively Progressive Bayesian Neural Network (\VIPARR) is aimed at pruning and growing individual layers of a single BNN to address domain incremental and class incremental scenarios of continual learning.

\subsection{Pruning the Hidden Layer of a Bayesian Neural Network}
As the weights of the BNN are probability distributions (approximated through Gaussian) instead of point estimates, the significance of the weight parameters is estimated through the signal to noise ratio, $SNR$ of the Bayesian weights defined as:
\begin{equation}
    SNR(w_{si}^k) = \frac{|\mu|_{si}^k}{\sigma_{si}^k};~~s=1,\ldots,K_o;~~ i = 1,\ldots,K_{o-1} \label{eqsnr}
\end{equation}
where $w_{si}^k$ refers to weight in layer k, linking from neuron $i$ in layer $k-1$ to neuron $s$ in layer $k$. It is to be noted that the higher the $SNR(.)$ of a weight, the larger its significance, and vice-versa.
\begin{figure*}
    \centering
    \includegraphics[width=\textwidth]{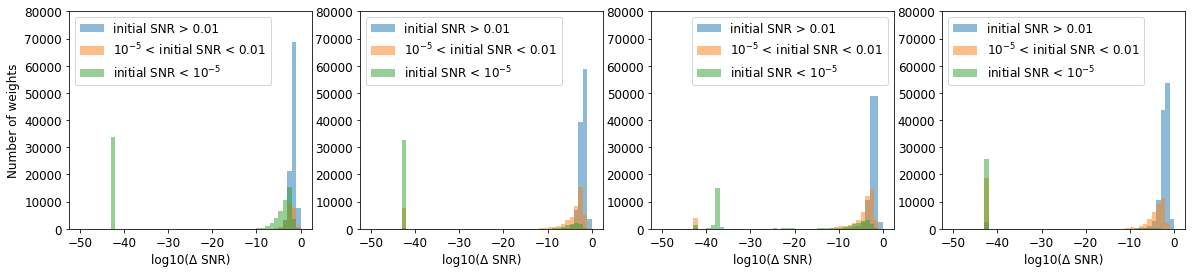} \\
     \caption{Statistics of Weight Adaptations across Tasks (left to right). The weights initialized with larger SNR (blue bars) are adapted more with the sequential tasks, and weights with smaller initial SNR (green bars) are adapted lesser as the network learns more tasks.}
    \label{fig:vcl_weightchange}
\end{figure*}

We first demonstrate the need for pruning the BNN while learning a sequence of tasks, using the split MNIST data set described in Section \ref{sec:results}. We train a BNN with 2 hidden layers, each with 256 neurons, for this demonstration. Then, we show the statistics of the redundant weight parameters in the BNN. We review the statistics of weight adaptations in the BNN, as shown in Fig. \ref{fig:vcl_weightchange} \cite{vcl}. Depending on the SNR at the end of training for task 1, we plot the histogram of the statistics of {\it weight adaptations} after each task, by sorting them into three bins, namely, (a) weights with initial $SNR(\mathbf{w}) >0.01$ (depicted in blue in Fig. \ref{fig:vcl_weightchange}), (b) weights with initial $10^{-5}<SNR(\mathbf{w})<0.01$ (depicted in orange in Fig. \ref{fig:vcl_weightchange}) and (c) weights with initial $SNR(\mathbf{w}) < 10^{-5}$ (depicted in green in Fig. \ref{fig:vcl_weightchange}). For ease of observation, we present them in a logarithmic 10 scale.

From the figure, it can be observed that the weights initialized with larger SNR (blue bars on Fig. \ref{fig:vcl_weightchange}) are adapted more with the sequence of tasks, and weights with smaller initial SNR (green bars on Fig. \ref{fig:vcl_weightchange}) are adapted lesser as the network learns more tasks. As the weights with smaller SNR do not participate in learning any task, they are insignificant and are pruned from the BNN. Therefore, after training the network for the task $t-1$, we identify the weights with $SNR(w_{si}^k) < \beta_k;~k=1,\ldots,o$, where $\beta_k$ is a user-defined threshold for insignificant weights in individual hidden layers of the BNN. We refer to the number of insignificant network connections pruned in each layer as $\delta_k$. These insignificant weights and their prior distributions are re-initialized (with a random normal distribution) before training the network for the task $t$.  

\subsection{Growing Individual Layers of a BNN}
Although the re-initialization of the pruned weights help the neurons in the BNN to adapt for the task $t$, we hypothesize that these weights alone are insufficient for representing the task $t$, as there are task specific representations that need additional resources. Therefore, we {\bf add sufficient neurons} to individual layers of the BNN for robust representation of all tasks. The number of neurons to be added for the task $t$, in each layer of the BNN towards robust continual learning is based on three factors:
\begin{itemize}
    \item Estimated required network resources.
    \item Estimated available network resources through shared representations with the previous tasks.
    \item Estimated available network resources for task-specific representation.
\end{itemize}

{\bf Estimated Required Network Resources ($\alpha_{req}^k$):} The number of neurons to be added to individual layer of the network is estimated to ensure that the network resources are fairly distributed across all tasks inspired by \cite{cl_progressivedeep}. Specifically, if the network starts with $n_{init}^{k}$ neurons on the kth layer for learning the first task, $\alpha_{req}^k$ would be $n_{init}^{k}$ for other tasks with the same number of classes and same input size. Please note that $\alpha_{req}^k$ could be different from the $n_{init}^{k}$ for more difficult or easier tasks with different input size and number of classes compared to the first task.

{\bf Estimated available network resources through shared representations with the previous tasks ($\alpha_{share}^k$):}
We estimate the shared representations for task $t$ with the preceding $t-1$ tasks through the mean activations of individual neurons for samples available in the classes considered. Thereafter, the average of all pairwise distances between the activations for the $tc$ classes are computed (Eq.\eqref{eq:activated_nn}). If the distance is above a user-defined threshold of $\gamma_k$, the neuron is known to contribute to the classification, and its representation is considered as an useful shared representation. 
\begin{equation}\label{eq:activated_nn}
    \alpha_{share}^k = \frac{2 \times \sum_{i=0}^{tc-1}\sum_{j>i}^{tc} dist(\phi_i, \phi_j)}{tc(tc-1)} > \gamma_k
\end{equation}
where $tc$ is the total number of classes in the output layer and $\phi_i$ is the mean activation of class $i$ in a certain neuron. The number of neurons that are estimated to contribute to classification of task $t$ is denoted by $\alpha_{share}^k$. These neurons share representations for the task $t$ with the previous tasks.

{\bf Estimated available network resources for task-specific representation ($\alpha_{prune}^{k}$):} The available network resources to represent task $t$ are estimated based on the number of weights pruned and re-initialized, at the end of training for task $t-1$. The number of pruned neurons in the $k^{th}$ layer of the network ($\alpha_{prune}^{k}$) is estimated based on:
\begin{equation}\label{eq:pruned}
    \alpha_{prune}^{k} \propto \frac{\delta_k}{size(k-1)} 
\end{equation}
where $\delta_k$ is the number insignificant weight parameter and $size(k-1)$ is the number of neurons in the layer $k-1$ that helps to estimate the number of neurons based on the number of pruned weight connections ($\delta_k$).

Thus, the number of neurons to be added to layer $k$ ($\alpha^k$) in the network for the task $t$ is estimated through \begin{equation}\label{eq:grow}
    \alpha^k = \alpha_{req}^k  - \left(\alpha_{share}^k + \alpha_{prune}^k\right)
\end{equation}
where $\alpha_{req}^k$ is the number of neurons required by the new task, $\alpha_{share}^k$ (Eq. \eqref{eq:activated_nn}) is the number of neurons that are being activated for the new task in the current network and $\alpha_{prune}^k$ (Eq. \eqref{eq:pruned}) is an estimated number of pruned neurons.

\subsubsection{Class Incremental Learning in \VIPARR}\label{plasoplayer}
As the number of tasks increases in a class incremental scenario, there is a need to increase the number of neurons in the output layer of the network. This can be done in either a multi-head setting or a single head setting. The {\it multi-head setting} allows for a task-specific output layer for each task, while in the {\it single head setting}, the output layer is expanded based on the number of classes in the new task. As the multi-head setting involves replacing the output layer with an output layer of the new classes, it is straightforward. On the other hand, we have specific strategies for the single-head setting. The most straightforward way to expand the output layer is to add new neurons, which are also initialized with random normal distributions, while ensuring that the weights of the existing output neurons are preserved. However, this causes bias, which is discussed below alongwith their mitigation strategies:
\begin{itemize}
    \item {\bf Backpropagation errors due to varied training status of output neurons:} The weights connecting output neurons of classes in tasks $1,\ldots,t-1$ have been trained for these classes at the end of task $t-1$. On the other hand, the new output neurons for classes in task $t$ are just initializations, and hence, this difference in training status of weights in the output layer at the beginning of training for task $t$ could cause misclassification. This could cause errors in backpropagation, resulting in erroneous weight adaptations \cite{cl_single_incremental_task_scenarios}, thus causing the weights to converge sub-optimally. Hence, to prevent this from occurring, all the weights in the output layer are re-initialized before training for task $t$. 
    \item {\bf Bias due to data imbalance caused by task progression:} As the neurons in the output layer are appended based on the data for task $t$, there is a data imbalance on distinct neurons due to the following: (a) existing neurons are oblivious to the data for task $t$, and (b) the neurons added for classes in task $t$ are oblivious to data from tasks $1,\ldots,t-1$. Therefore, with the re-initialization of weights, weights adapted after training for task $t$ are biased towards the classes in the task $t$, because the loss is calculated only based on the data from these classes. Hence, replacing the output weights corresponding to the classes in the previous tasks, with weights obtained after replay for task $t-1$ can help to alleviate (b) above. Thereafter, (a) can be alleviated through replay after training for task $t$. This would further help to finetune the output weights for robust predictions.
\end{itemize}

\subsection{Additional Strategies to avoid Catastrophic Forgetting}
In addition to the pruning and growing of the Bayesian neural network, we also use {\it adaptive regularization} of the variational weights across multiple tasks \cite{vcl} to ensure that the representations of the past task are not forgotten. Therefore, the variational weights $\mathbf{w}$ parametrized by $\theta$ of the BNN for task $t$  are estimated through minimizing the loss function, which consists of the cross entropy loss and a $KL$-divergence regularization term, as shown in Eq. \eqref{fullloss}.
\begin{eqnarray}
    \mathcal{L}^t(q_t(\theta))  = \sum_{n=1}^{N_t}\mathbb{E}_{\theta~q_t(\theta)}[log~p(y_t^{(n)}|\theta, x_t^{(n)})]\nonumber\\ - KL(q_t(\theta)||q_{t-1}(\theta)) \label{fullloss}
\end{eqnarray}
The KL-divergence measures the difference between the posterior distribution of the current task $q_t(\theta)$ and its prior distribution. This prior distribution is typically the posterior distribution of the network at the end of the previous task $q_{t-1}(\theta)$. In the case of the first task, the prior distribution $q_0(\theta)$ is given by a zero-centered Gaussian distribution. It must be noted that this regularization prevents large changes in the weight parameters to preserve representations of tasks $1,\ldots, t-1$. 

Moreover, as the BNN is being adapted for the task $t$, {\it replaying a few samples from previous tasks} can help prevent catastrophic forgetting, especially, to introduce samples from past task to the new neurons that are progressively added in each layer of the network. In addition, it allows the newly added neurons to be trained on a subset of data from the old tasks \cite{GEM,icarl}. To this end, we build a coreset ($CT$), which is a collection of randomly selected subset of samples from each task. Depending on whether the class incremental scenario is addressed using multi-head or single-head, the replay is performed as under:
\begin{itemize}
    \item Multi-head setting: Each task has a unique output layer, corresponding to the number of output classes in each task. This calls for a unique task identifier during the training process and inference. The inferences are drawn using the output layer corresponding to the task label from which the sample originated. Hence, it must be noted that the task labels are required at the time of inference in the multi-head setting. As this entails task specific neurons in the output layer, the coreset of the individual task is only replayed during inference for that task.
    \item Single head setting: Alternatively, in  a single head setting, the number of neurons in the output layer are also increased according to the number of classes in the new task. Thus, in this setting, all the layers of the network are shared across multiple tasks, and this is agnostic to tasks. The model is retrained using the entire coreset $CT = [CT_{1}\ldots CT_{t-1} CT_{t}]$ at the end of each task $t$ training. It must be noted that as the single head setting is task agnostic, it is not required to specify task labels at inference. 
\end{itemize} 

The algorithm of the \VIPARR~is summarized in Algorithm \ref{cebancoalgo}. 

\begin{algorithm}[h]
\SetAlgoLined
Initialize the network; $\beta_k, k = 1,\ldots,o$; $\gamma_k, k = 1,\ldots,o$; Weights of the network ($\mathbf{w}_k; k = 1,\ldots,o$)\\
\KwIn{Dataset for task 1: $\mathcal{D}_{1}$ \\
Randomly sample coreset for task 1 ($CT_1$)\\
Update the variational distribution of weights using $\left(\mathcal{D}_{1} \setminus CT_{1}\right)$ with Eq. \eqref{fullloss}.\\
\For{$t=2,\ldots,l$}{
 Append the coreset ($CT$) with randomly sampled $CT_{t}$ from $\mathcal{D}_{t}$. \\ 
 Re-initialize less significant weights in existing network. \\
 Grow hidden layers according to Eq. \eqref{eq:grow}.\\
 \If{single head}{
  Grow output layer by number of new classes.\\
  Update the variational distribution of weights using \\($\mathcal{D}_{t} \setminus CT_{t}$) with Eq. \eqref{fullloss}.\\
  Update the variational distribution of weights with $CT = \left[CT_{1}\ldots CT_{t}\right]$.\\
  }  
  \If{Multi-head}{
  Update the variational distribution of weights using \\($\mathcal{D}_{t} \setminus CT_{t}$) with Eq. \eqref{fullloss}.
  }
  }
 }
Evaluation Phase:\\
\For{$t=1,\ldots,l$}{
\If{multi-head}{
Fit the relevant task $t$ head. \\
Update the variational distribution of weights with $CT_{t}$.}
Predict using evaluation data of task $t$. }
 \caption{Learning Algorithm of \VIPARR}\label{cebancoalgo}
\end{algorithm}

Next, we demonstrate the effect of the individual strategies, and the effectiveness of the proposed approach on a number of data sets.

\section{Performance Evaluation} \label{sec:results}
In this section, we evaluate the effectiveness of the individual strategies of the comprehensively Progressive BNN for robust continual learning \VIPARR, using the sequence of task settings, as listed in Table \ref{tab:expsetting}. From this table, it can be observed that the class incremental representational ability of the \VIPARR~is evaluated in both the multi-head and single head settings.

\begin{table}[ht]
\centering
\caption{Experimental Settings for Evaluation. MH refers to the multi-head setting and TA refers to task-agnostic and also the single head setting. }\label{tab:expsetting}
\begin{tabular}{|c|c|c|c|c|c|}\hline
Scenario & \multicolumn{2}{|c|}{Data} & Setting & Number & Number of \\\cline{2-3}
& Data set & Image Size & & of Tasks&Classes per task\\\hline
Homo DI & \multirow{3}{*}{MNIST} &\multirow{3}{*}{28$\times$ 28} & pMNIST & 10 & 10\\\cline{1-1}\cline{4-6}
Homo TI & \multirow{3}{*}{} &\multirow{3}{*}{} & MH-sMNIST & 5 & 2\\\cline{1-1}\cline{4-6}
Homo CI & \multirow{3}{*}{} &\multirow{3}{*}{} & TA-sMNIST & 5 & 2\\\hline
Homo TI & \multirow{2}{*}{CIFAR 100} &\multirow{2}{*}{32$\times$32$\times$3} & MH-CIFAR100 & 20 & 5\\\cline{1-1}\cline{4-6}
Homo CI & \multirow{2}{*}{} &\multirow{2}{*}{} & TA-CIFAR100& 20 & 5\\\hline
\multirow{3}{*}{Hetero TI} & MNIST & $28 \times 28$ & \multirow{3}{*}{}& \multirow{3}{*}{3} & \multirow{3}{*}{10}\\\cline{2-3}
\multirow{3}{*}{} & SVHN & $32 \times 32$ & \multirow{3}{*}{}& \multirow{3}{*}{} & \multirow{3}{*}{}\\\cline{2-3}
\multirow{3}{*}{} & CIFAR10 & $32 \times 32 \times 3$ & \multirow{3}{*}{}& \multirow{3}{*}{} & \multirow{3}{*}{}\\\hline
\end{tabular}
\end{table}

In all our experiments, we use average task accuracy (Accuracy) for principled evaluation. Let us assume that all the test data of $l$ tasks are available for evaluation and the accuracy of task $t$ after training on $l$ tasks is $A_{lt}$. Then the average task accuracy is given by:
\begin{equation}
   Accuracy =  \frac{1}{l} \sum_{t = 1}^{l} A_{lt} \label{acc}
\end{equation}
It must be noted that the two hyperparameters of \VIPARR, namely, the user defined threshold for insignificant weights ($\beta_k$) and user-defined threshold for average of pair-wise distance between activations ($\gamma_k$) are initialized with the same value for all $k=1,\dots,o$ layers in this study. Hence the $\beta_k$ and $\gamma_k$ can be replaced with $\beta$ and $\gamma$.

The SOTA methods used in comparison with \VIPARR~are the Synaptic Intelligence (SI) \cite{si}, Progressive Neural Networks (PNN) \cite{PNN}, elastic weight consolidation (EWC) \cite{EWC2017}, Incremental Classifier and Representation Learning (iCARL) \cite{icarl}, Gradient Episodic Memory (GEM) \cite{GEM}, Riemannian Walk (RWalk) \cite{rwalk}, and the variational continual learning (VCL) \cite{vcl} approaches, using the MNIST and CIFAR100 data sets. We also report the average evolved network structure, along with the performance metrics (Eq. \eqref{acc}), over 5 validations with different random seeds. 

\subsection{Demonstration of \VIPARR~on MNIST:}
First, we present the results of \VIPARR~on the MNIST data set under the three different scenarios, viz., permuted MNIST (pMNIST), multi-head split MNIST (MH-sMNIST) and Task Agnostic (single-head) split MNIST (TA-sMNIST) in Table \ref{tab:mnist_performance}. It can observed that the \VIPARR~outperforms most SOTA methods in continual learning and has comparable accuracy with PNN. However, it should be noted that the PNN is a growing network that expands by [256,256] for each task. Therefore, the final network size for PNN is [2560, 2560]. The \VIPARR~converges with accuracies similar to that of PNN, with fewer network resources. In the task agnostic class incremental scenario of split MNIST data set (TA-sMNIST), the strategies of \VIPARR~helps to improve the accuracy of Bayesian neural network for continual learning from $60.6\%$ to $83.8\%$, improving by a significant $23.2\%$. Overall, \VIPARR~outperforms other SOTA methods (RWalk) by at least $1.3\%$ in this scenario. It is also observable that \VIPARR~starts with a minimal architecture, and evolves with pruning the weights and growing as required only. Thus, \VIPARR~is capable of representing distributional and class increments efficiently with a compact network architecture. 

\begin{table}
\begin{center}
\renewcommand{\arraystretch}{1.05}
\caption{Performance Results of \VIPARR~ on MNIST Dataset} \label{tab:mnist_performance}
\begin{tabular}{clccc}
\hline
\hline
\textbf{Dataset} & \textbf{Methods}     & \textbf{Initial}  & \textbf{Final} & \textbf{Accuracy} \\
&                     & \textbf{Network}  & \textbf{Network} & \textbf{(\%)} \\\hline
&EWC     & $[100, 100]$        & $[100, 100]$         & $84.0$ \cite{vcl}   \\\cline{2-5}
&SI      & $[100, 100]$        & $[100, 100]$        & $86.0$ \cite{vcl}   \\\cline{2-5}
&LP      & $[256, 256]$        & $[256, 256]$        & $82.0$ \cite{vcl}  \\\cline{2-5}
&GEM     & $[256, 256]$        & $[256, 256]$        & $93.1$ \cite{chaudhry2018efficient}     \\\cline{2-5}
pMNIST&RWalk  & $[256, 256]$         & $[256, 256]$        & $91.6$ \cite{chaudhry2018efficient}    \\\cline{2-5}
&PNN    & $[256, 256]$         & $[2560, 2560]$        & $94.6$ \cite{chaudhry2018efficient}    \\\cline{2-5}
&VCL    & $[100, 100]$         & $[100, 100]$          & $93.0$ \cite{vcl}   \\\cline{2-5}
&\VIPARR   & $[32, 32]$        & $[58, 93]^{*}$     & \textbf{$92.7 \pm 0.14$}  \\\cline{2-5}
&\VIPARR   & $[32, 32]$        & $[126, 139]^{*}$     & \textbf{$94.2 \pm 0.23$} \\\hline

&EWC     & $[256, 256]$        & $[256, 256]$      & $63.1$ \cite{vcl}  \\\cline{2-5}
&SI      & $[256, 256]$        & $[256, 256]$      & $98.9$ \cite{vcl} \\\cline{2-5}
&LP      & $[256, 256]$        & $[256, 256]$      & $61.2$ \cite{vcl}  \\\cline{2-5}
&GEM     & $[256, 256]$        & $[256, 256]$      & $94.3$ \cite{ebrahimi2020adversarial}    \\\cline{2-5}
MH-sMNIST&RWalk  & $[256, 256]$         & $[256, 256]$      & $99.3$ \cite{rwalk}    \\\cline{2-5}
&iCaRL  & $[256, 256]$         & $[256, 256]$      & $89.3$ \cite{ebrahimi2020adversarial}    \\\cline{2-5}
&PNN    & $[256, 256]$         & $[1280, 1280]$      & $99.8 \pm 0.05$     \\\cline{2-5}
&VCL    & $[256, 256]$         & $[256, 256]$        & $98.4$ \cite{vcl}  \\\cline{2-5}
&\VIPARR   &$[64, 64]$         & $[116, 112]$     & $99.3 \pm 0.13$  \\\hline

&EWC     & $[256, 256]$       & $[256, 256]$   & $55.8$ \cite{rwalk}   \\\cline{2-5}
&RWalk  & $[256, 256]$        & $[256, 256]$     & $82.5$ \cite{rwalk}   \\\cline{2-5}
TA-sMNIST&iCaRL  & $[256, 256]$       & $[256, 256]$     & $55.8$ \cite{rwalk}   \\\cline{2-5}
&VCL    & $[160, 128]$        & $[160, 128]$    & $60.6 \pm 4.41$  \\\cline{2-5}
&\VIPARR     & $[128, 128]$  & $[160, 128]$    & \textbf{$83.8 \pm 0.38$} \\\hline
\end{tabular}
\end{center}
\begin{tablenotes}
\item $*$ With different threshold $\gamma$ for average pair-wise distances among activations from different classes at neuron level, we are able to get different final network size and accuracy using \VIPARR.
\end{tablenotes}
\end{table}
Next, we present the effectiveness of the proposed strategies, followed by a study on the effect of hyperparameters. All these studies are based on the task agnostic split MNIST (TA-SMNIST) scenario.

\vspace{0.4cm}
\subsubsection{Study on Effectiveness of the proposed \VIPARR:} 
\begin{figure*}[h]
\centering 
\captionsetup{justification=centering}
\subfigure[Mean Activations of Neurons in Hidden Layer 1: Tasks 1, 2 and 3]{\includegraphics[width=\textwidth,height=0.25\textheight,valign=t]{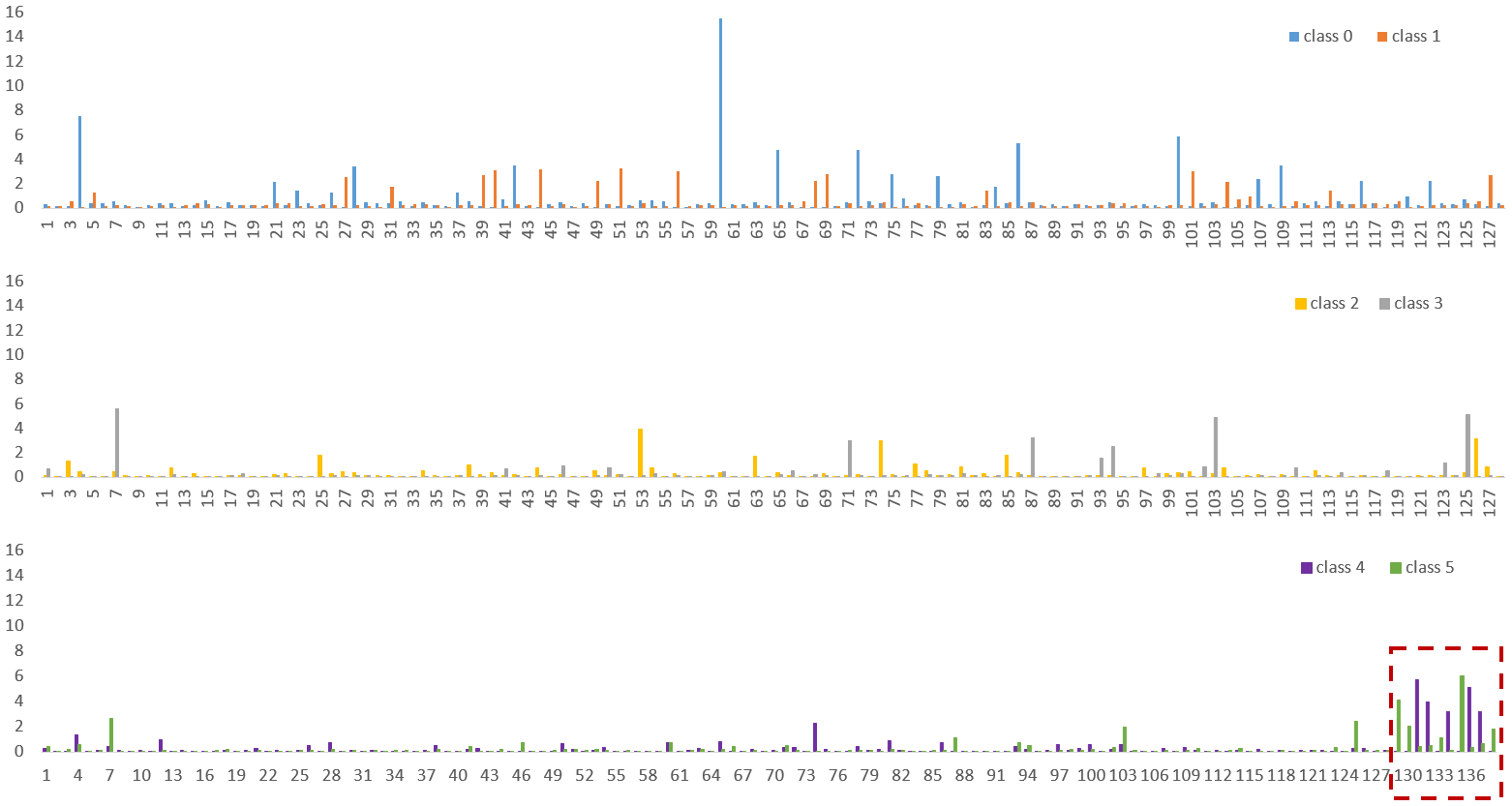}\label{firstlayer}}
\subfigure[Mean Activations of Neurons in Hidden Layer 2: Tasks 1, 2 and 3]{\includegraphics[width=\textwidth,height=0.25\textheight,valign=t]{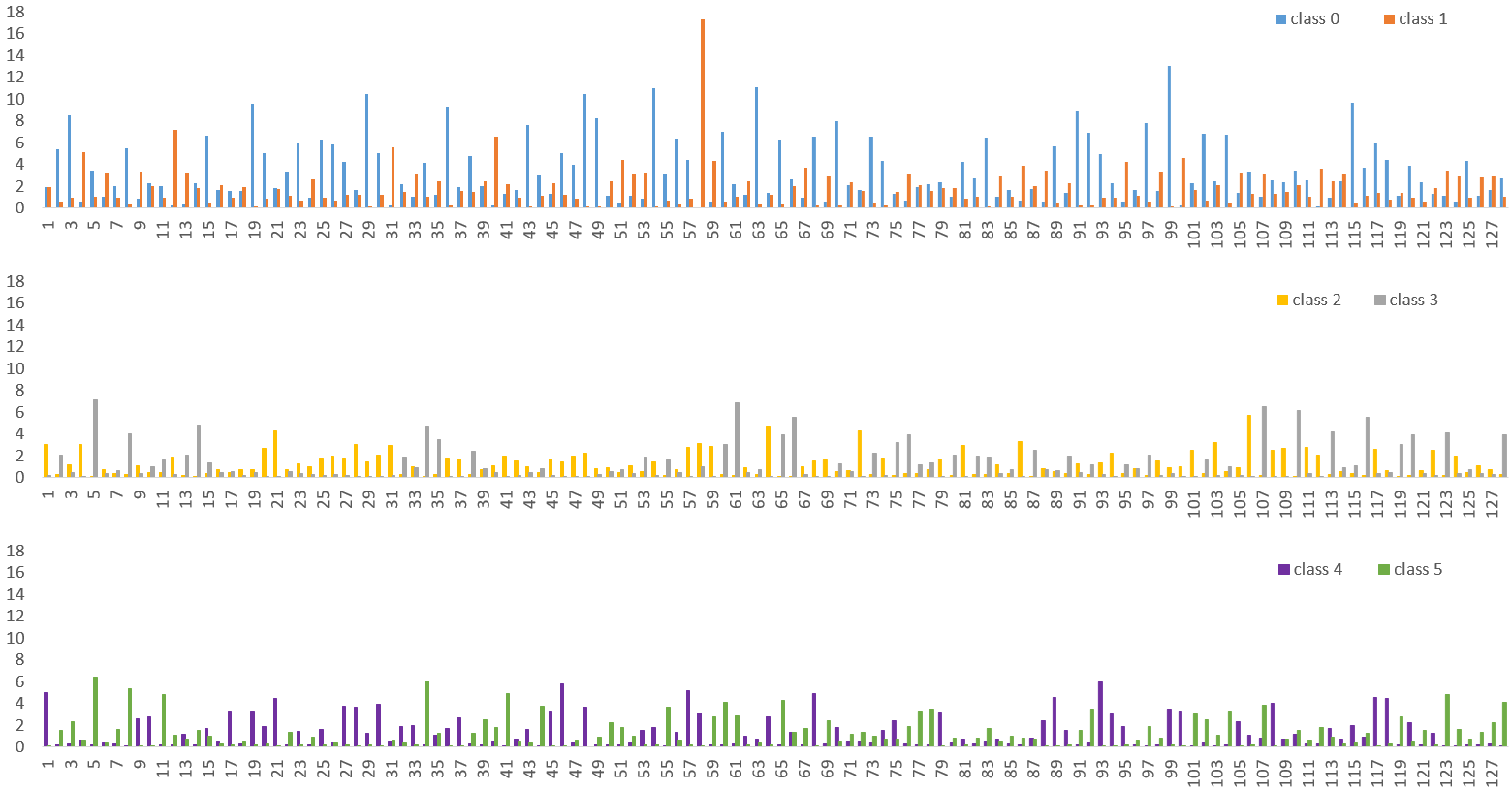}\label{seclayer}}
\caption{Mean Activations of Neurons in Hidden Layers for Samples in Classes of individual Tasks. Blue: class 0; Orange: class 1; Yellow: class 2; Grey: class 3; Purple: class 4; Green: class 5.}
    \label{fig:avtivations}
\end{figure*}

First, we show that the proposed comprehensively progressive BNN enables a parsimonious network structure. This is because the strategies add neurons only when it is absolutely essential. We demonstrate this through presenting the activations of the various neurons in both the hidden layers of the BNN, for the classes in the new task, while training for the TA-sMNIST scenario in Fig. \ref{fig:avtivations}. From the figure, it can be observed that the neurons in hidden layer 2 are capable of discriminating samples of each class in the individual new tasks, and hence, no neurons are added to the hidden layer 2 of the network. Similarly, as the existing neurons in hidden layer 1 are capable of distinguishing classes in Task 1, no neurons are added to the layer. However, the fewer neurons (inset of Fig. \ref{firstlayer}) are added while Task 2 is introduced. 

\begin{figure*}[htp]
    \centering
    \includegraphics[width=\textwidth]{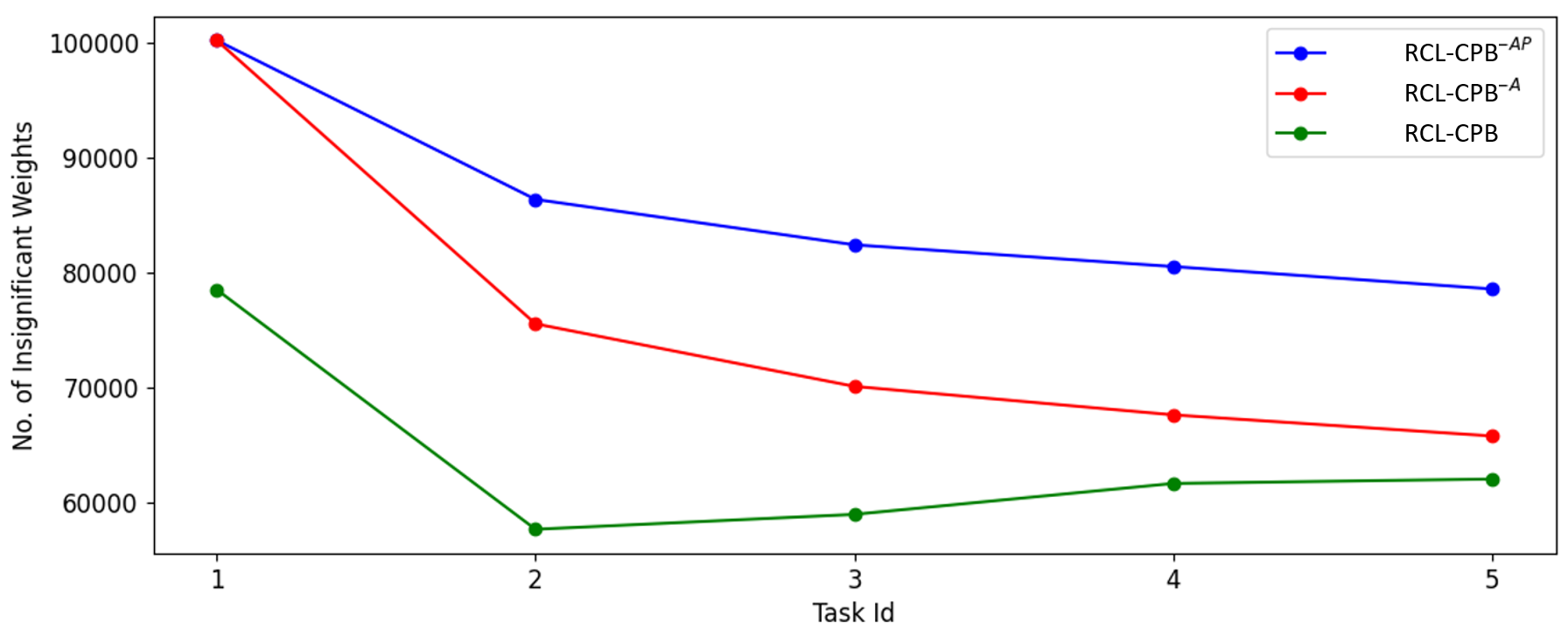}
    \caption{Effect of pruning and neuron addition strategies in the comprehensively Progressive Bayesian Neural Network on the Utilization of the Network Resources}
    \label{fig:vip_growing_results}
\end{figure*}

Next, Fig. \ref{fig:vip_growing_results} presents the statistics of significant weight parameters in a network without neuron addition (\VIPARR$^{-A}$) and network without addition and pruning (\VIPARR$^{-AP}$). To ensure fair comparison, all these methods are based on a two layer network with $[160,128]$ neurons (the final size of network \VIPARR~converged to). From the figure, it can be seen that the \VIPARR~has a better utilization of the network resources. It can also be observed that pruning with re-initialization (\VIPARR$^{-A}$) helps to improve network utility. It is also evident that a large network at the beginning of learning is not essential, as the network utility is poorer (with many redundant resources) for training tasks 0 and 1. Thus, it can be observed from Table \ref{tab:mnist_performance} and Fig. \ref{fig:vip_growing_results} that the strategies for structural adaptations help to improve accuracy, while ensuring improved utility of network resources.

\begin{figure*}[htp]
\centering 
\captionsetup{justification=centering}
\subfigure[\VIPARR$^{-AP}$]{\includegraphics[width=\textwidth,height=0.25\textheight]{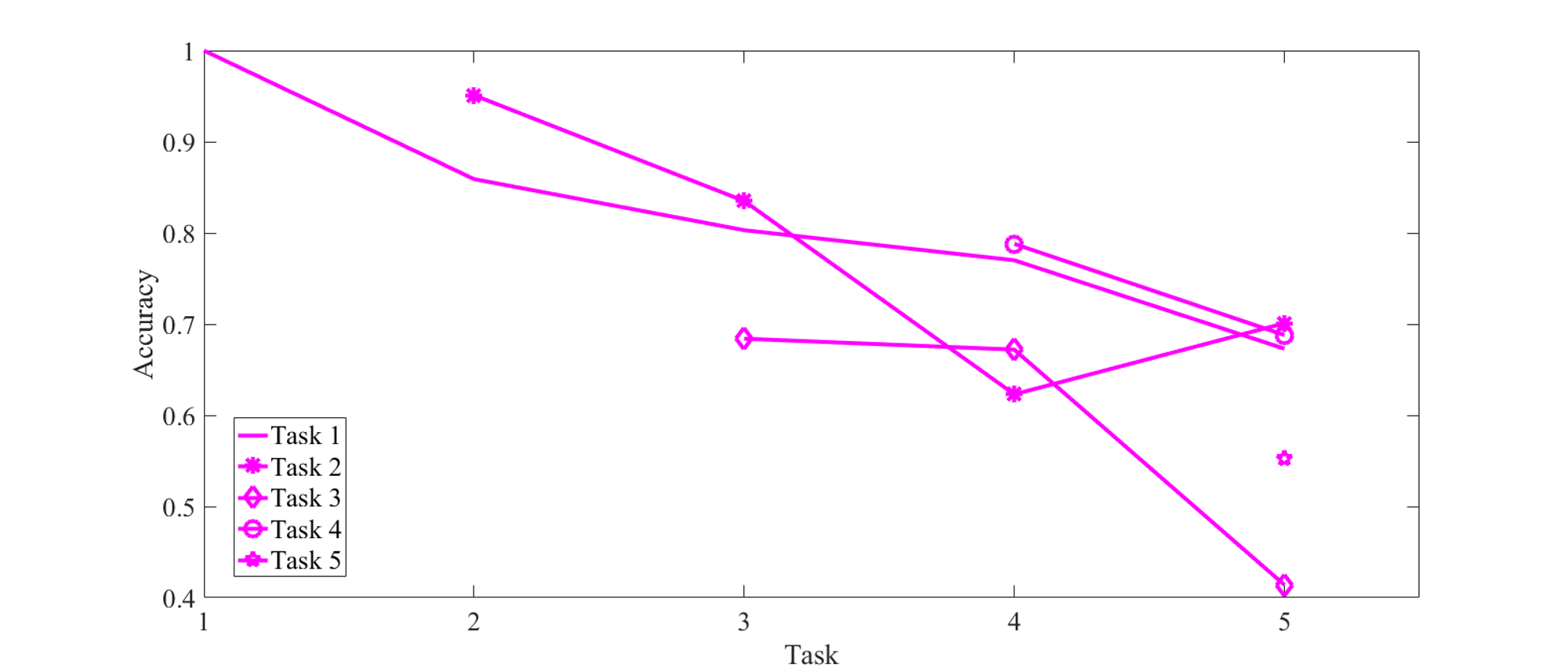}\label{catforap}}
\subfigure[\VIPARR$^{-A}$]{\includegraphics[width=\textwidth,height=0.25\textheight]{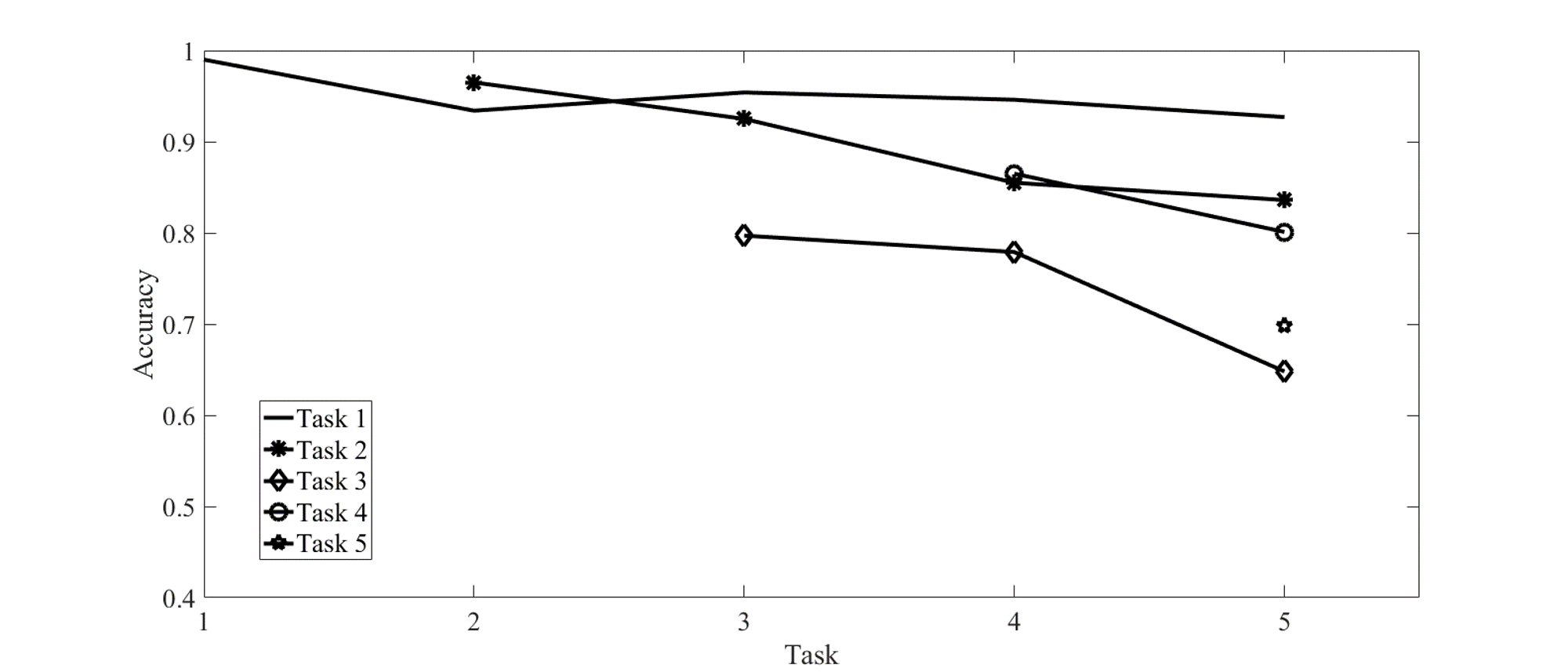}\label{catforp}}
\subfigure[\VIPARR]{\includegraphics[width=\textwidth,height=0.25\textheight]{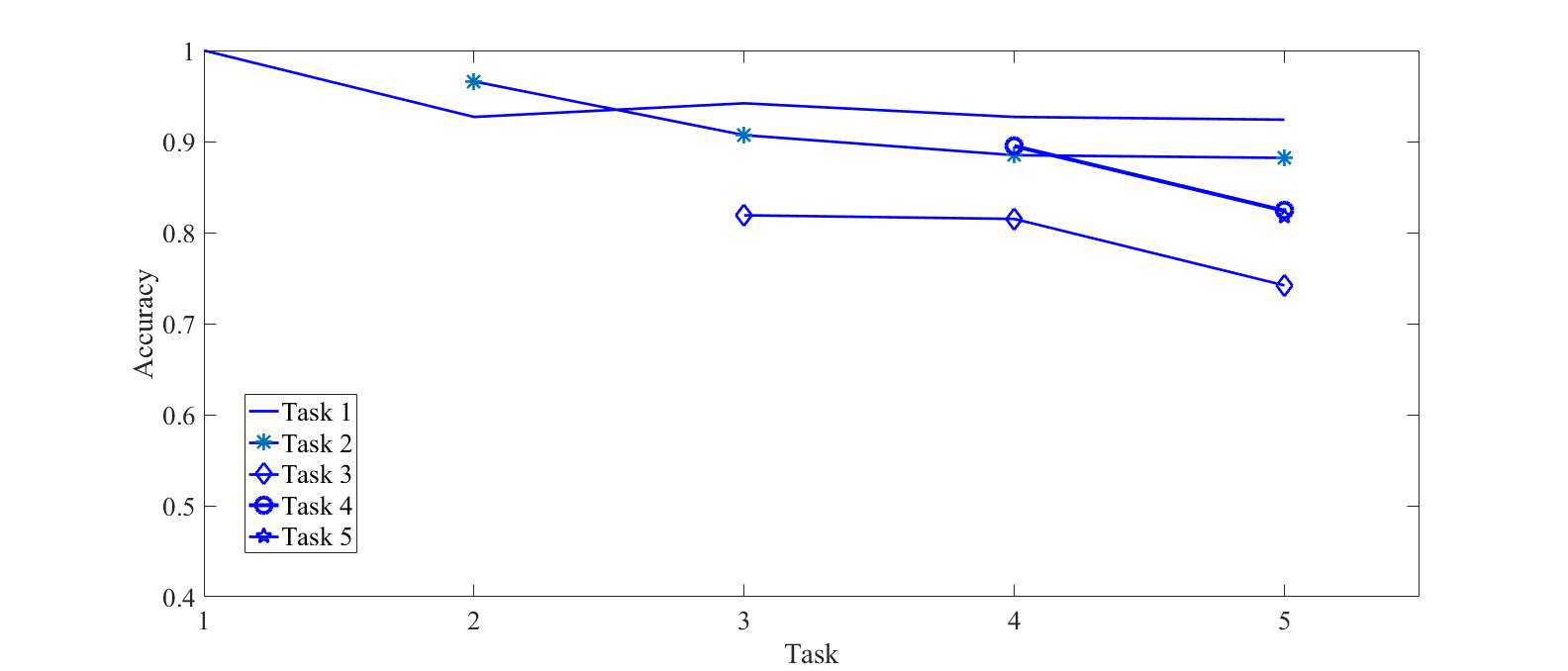}\label{catfor}}
\caption{Effect of pruning and neuron addition strategies in the comprehensively Progressive Bayesian Neural Network on Catastrophic Forgetting. Figure reports the accuracy of individual tasks ($1,\ldots, l$) after continually learning a sequence of tasks until task $l$. The drop in accuracy for a task ($1,\ldots, t-1$) after training on subsequent task $t$ is representative of the forgetting of that task. X-axis represents the number of sequential tasks $1,\ldots,l$ that the network has been trained on; Y-axis represents the testing accuracy for individual tasks; Legends represent the task that is being evaluated.}\label{fig:forgetting_singlehead}
\end{figure*}

Finally, we present the effectiveness of the structural adaptations in overcoming catastrophic forgetting. Fig. \ref{fig:forgetting_singlehead} shows the accuracy of the network for all classes in the preceding and the current task, at the end of training for each task. The accuracies of \VIPARR~is compared against \VIPARR$^{-A}$ and \VIPARR$^{-AP}$. From the figure, it can be observed that the strategies for structural adaptations in \VIPARR~help to overcome catastrophic forgetting, compared to \VIPARR$^{-A}$ and \VIPARR$^{-AP}$. This is especially evident from the accuracies of Task 3, for which the catastrophic forgetting of the network with neither of the strategies for structural intervention has a forgetting of $\approx30\%$, the network with pruning alone has a forgetting of $\approx15\%$, and the network with both the strategies of structural intervention has very minimal forgetting (<8\%). It must also be noted that the initial accuracy for the task is also higher with all the strategies. In general, the improved strategies for progressive BNN help to remember past tasks, while representing new tasks accurately.

\subsubsection{Effect of the hyperparameters in the strategies on the \VIPARR:}

\begin{figure*}[htp]
\centering 
\captionsetup{justification=centering}
\subfigure[Effect of $\gamma$ on Accuracy]{\includegraphics[width=0.49\textwidth]{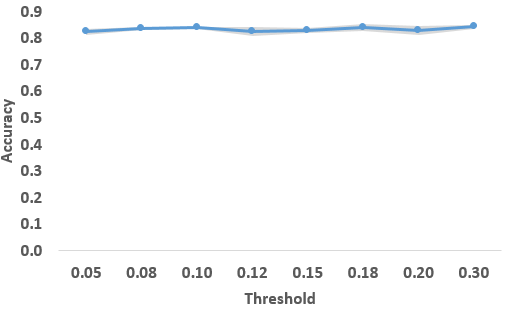}\label{coreset_permuted}}
\subfigure[Effect of $\gamma$ on Network Size]{\includegraphics[width=0.49\textwidth]{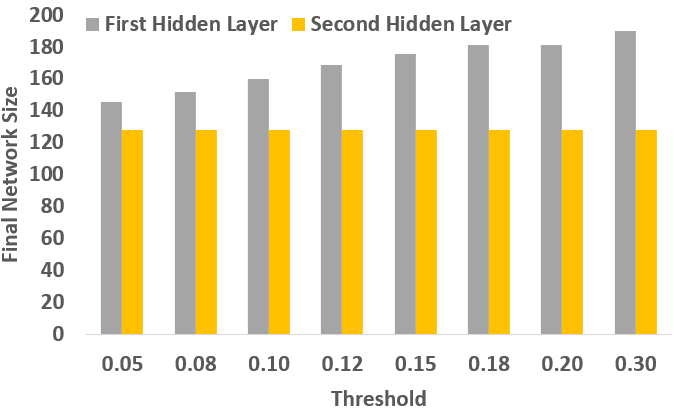}\label{coreset_split}}
\caption{Effect of $\gamma$}\label{fig:gammaeffect}
\end{figure*}

We study the effect of hyperparameters, $\gamma$ and the size of the initial network, to the final network size and the accuracy of the network. It must be noted that $\gamma$ is the threshold of the pairwise distances between activations of all the classes. It is used to estimate the number of neurons in the individual layer of the networks that can be shared across tasks. In all our experiments, we set the same $\gamma$ for all hidden layers in the network. From Fig. \ref{fig:gammaeffect}, which is a study on the effect of $\gamma$, it can be observed that the accuracy is less affected by the threshold $\gamma \in (0.05,0.3)$. However, the size of the network increases with increasing $\gamma$, especially, in the first hidden layer. Thus, a threshold of $0.05<\gamma<0.3$ is a suitable range to estimate shared representations across tasks.

\begin{figure*}[htp]
\centering 
\captionsetup{justification=centering}
\subfigure[Effect of Initial Size on Accuracy]{\includegraphics[width=0.49\textwidth]{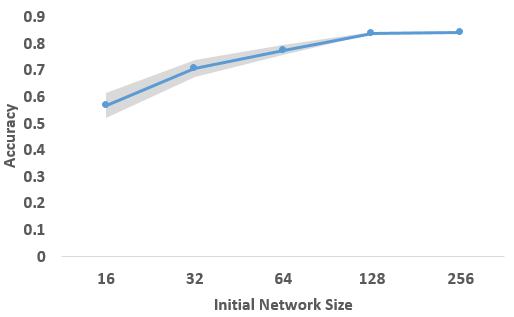}\label{coreset_permuted}}
\subfigure[Effect of Initial Size on Network Size]{\includegraphics[width=0.49\textwidth]{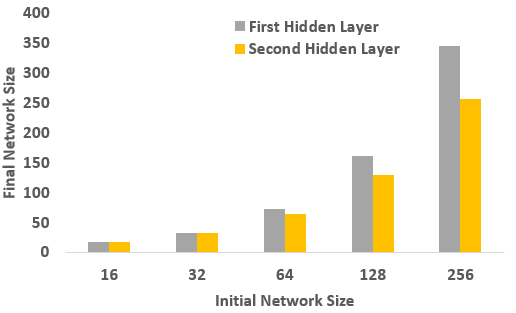}\label{coreset_split}}
\caption{Effect of Initial Size of the Network}\label{fig:initnwsize}
\end{figure*}

Fig. \ref{fig:initnwsize} presents the results of our study on the effect of the initial size of the network. We vary the initial size of the network between [16,16] to [256,256] and report the average accuracy for each network size, over 5 runs. From the figure, it can be observed that the average accuracy over the 5 tasks and the final network size increases with increasing initial network size. However, the size of the network is quite large for an initial network size of [256,256], while the gain in accuracy is very minimal. Hence, we choose an initial network size of [128,128], for the TA-sMNIST scenario of continual learning.

From the results in the section, it can be observed that the structural adaptations and the regularization helps to enhance the perception and memory of BNN. As the effect of the recollection intervention strategy has already been explored in \cite{vcl}, which shows that the performance improves with the increasing coreset size although it asymptotes for large coresets, it is not demonstrated again in detail here. Following these findings, we evaluate the performance of the \VIPARR~on more complex data sets, viz., CIFAR100 and the sequence of heterogeneous tasks.

\subsection{Performance Studies on CIFAR-100}
The results for CIFAR-100 on multi-head is presented in Table \ref{tab:cifar100_multihead_performance}. Our experiments on \VIPARR$^{-AP}$~(which is the same as VCL \cite{vcl}) and \VIPARR~are conducted using a BNN with 2 convolutional layers (of size 16 and 32, each, with a filter of 3x3), followed by two fully connected layers. While the \vcl~has a fixed network size, the fully connected layers of \VIPARR~are pruned and grown during the learning. From the Table \ref{tab:cifar100_multihead_performance}, it can be observed that the proposed \VIPARR~ improves average classification accuracy across all tasks by at least $2.3\%$ (compared to other SOTA algorithms). Especially on the variational inference based continual learning methods, the structural adaptations help to improve accuracy by $\approx 5\%$ in the multi-head setting.

\begin{table}
\begin{center}
\renewcommand{\arraystretch}{1.05}
\caption{CIFAR100 Multihead} \label{tab:cifar100_multihead_performance}
\begin{tabular}{lccc}
\hline
\hline
\textbf{Methods}      & \textbf{Initial network size}  & \textbf{Final network size}   & \textbf{Accuracy(\%)} \\ \hline
EWC     & $ResNet18^b$  & $ResNet18^b$        & $42.4$ \cite{chaudhry2018efficient}   \\ \hline
PI      & $ResNet18^b$  & $ResNet18^b$       & $47.1$ \cite{chaudhry2018efficient}   \\ \hline
GEM     & $ResNet18^b$  & $ResNet18^b$       & $65.4$ \cite{GEM}     \\ \hline
RWalk  & $ResNet18^b$   & $ResNet18^b$       & $70.1$ \cite{rwalk}     \\ \hline
iCaRL  & $ResNet18^b$   & $ResNet18^b$       & $50.8$ \cite{GEM}     \\ \hline
PNN    & $ResNet18^b$   & $ResNet18^b \times 20$       & $59.2$ \cite{chaudhry2018efficient}    \\ \hline
VCL    &  $[733, 789]^a$     &  $[733, 789]^a$   & $67.6 \pm 0.74$ \\ \hline
\VIPARR    &  $[128,128]^a$ & $[733, 789]^a$     & $72.4 \pm 0.40$  \\ \hline
\end{tabular}
\end{center}
\begin{tablenotes}
\item $^a$ BNN with 2 convolutional layers of size 16 and 32, followed by \VIPARR~ in FC layers
\item $^b$ a reduced version of $ResNet18$ \cite{he2016deep}
\end{tablenotes}
\end{table}

Table \ref{tab:cifar100_singlehead_performance} presents the results of the CIFAR100 in task agnostic class incremental setting, where the performance of \VIPARR~is compared with that of \vcl. It can be observed that the pruning and growing of the fully connected layers in the BNN helps to improve accuracy of learning the CIFAR100 data set continually by $\approx2\%$. We hypothesize that pruning and growing the convolutional layers dynamically can help to further improve performances and this would be explored in our future work.
\begin{table}
\begin{center}
\renewcommand{\arraystretch}{1.05}
\caption{CIFAR100 Singlehead} \label{tab:cifar100_singlehead_performance}
\begin{tabular}{lccc}
\hline
\hline
\textbf{Methods}     & \textbf{Initial network size}  & \textbf{Final network size}  & \textbf{Accuracy(\%)} \\ \hline
VCL     & $[463,292]$     & $[463,292]$    &$21.3 \pm 0.79$    \\ \hline
\VIPARR    & $[256, 256]$   &$[463,292]$      &$23.0 \pm 0.38$  \\ \hline
\end{tabular}
\end{center}
\end{table}

\subsection{Performance Studies on a Sequence of Heterogeneous Data Sets}
Table \ref{tab:sequence_performance} presents the results of the proposed \VIPARR~,in comparison to the VCL, on a sequence of data sets, viz., MNIST, SVHN and CIFAR 10, in a  multi-head output setting. The results for VCL is obtained with the final network structure that the \VIPARR~evolves to. It can be observed that the proposed \VIPARR~outperforms the VCL by $\approx6\%$. The architectural pruning and growing of the BNN helps with robust representation of the sequence of heterogeneous data sets.

\begin{table}
\begin{center}
\renewcommand{\arraystretch}{1.05}
\caption{Sequence of Tasks: MNIST -> SVHN -> CIFAR10} \label{tab:sequence_performance}
\begin{tabular}{lccc}
\hline
\hline
\textbf{Methods}     & \textbf{Initial network size}  & \textbf{Final network size}  & \textbf{Accuracy(\%)} \\ \hline
VCL     & $[636, 758]^d$           & $[636, 758]^d$        & $75.7 \pm 0.90$    \\ \hline
\VIPARR     & $[256, 256]^d$         & $[636, 758]^d$     & $81.5 \pm 0.13$  \\ \hline
\end{tabular}
\begin{tablenotes}
\item $^d$ Added with two convolutional layers with size 16 and 32 (filter size 3*3)
\end{tablenotes}
\end{center}
\end{table}

The studies in the section elucidates the following:
\begin{itemize}
\item The strategies for structural adaptation helps a Bayesian Neural Network to continually learn a sequence of tasks robustly in all scenarios of continual learning, viz., task incremental, domain incremental, class incremental, and sequence of heterogeneous tasks
\item The structural adaptation strategies enable improved utility of network resources, while ensuring that the representations are shared across tasks, wherever possible.
\item The improvement in accuracy is $>5\%$ for complex datasets. While the accuracies improve by at least $\approx2\%$ in the task agnostic class incremental scenario. 
\item The performance of the network is less sensitive to the hyperparameters of the strategies.
\end{itemize}

\section{Conclusion} \label{sec:conclusion}
This work presents a robust continual learning algorithm using a comprehensively progressive Bayesian Neural Network, through pruning of weights and addition of  neurons in individual layers of the network. The neuron addition is aimed at providing sufficient resources for individual tasks towards fair allocation of network resources, while ensuring shared representations of the network across tasks. The effects of the strategies and the effectiveness of the proposed method are demonstrated on the MNIST data set, under three different continual learning scenarios. Further to this, we evaluate the performance of \VIPARR~for learning a sequence of tasks continually, where tasks are defined based on a homogeneous data set (CIFAR-100 data set), and a heterogeneous sequence of tasks using MNIST, SVHN and CIFAR10 data sets. Performance results show that the proposed \VIPARR~is effective in addressing distribution incremental, class incremental and task agnostic class incremental scenarios. The improvement in performance is substantial in multi-head class incremental scenarios ($>5\%)$ for complex data sets. Although the accuracy improves by $\approx2\%$ in a task agnostic class incremental scenario, there is a need for specific strategies to enhance the performance of \VIPARR~ in this scenario. Thus, it can be observed from the demonstrations and the performance results that the proposed method improves the continual learning ability of BNN, and the proposed \VIPARR~ is robust to distributional and class incremental drifts. Moreover, the continual learning ability of the \VIPARR~is less sensitive to the hyperparameters of the learning strategies. Future work may also include structural adaptations of other architectures such as convolutional and recurrent Bayesian networks.

\section*{Acknowledgement}
The authors would like to thank the HBMS IAF-PP grant H19/01/a0/023 towards Diabetes Clinic of the Future Programme, and Institute for Infocomm Research, A*STAR, for supporting the study.
\bibliographystyle{elsarticle-num}
\bibliography{ref}

\end{document}


\maketitle
\section{Hyperparmameters}
In this section, hyperparameters used for different experiments are provided. In the MNIST demonstrations, a two layer MLP is used for 3 different scenarios on MNIST dataset and the hidden size for both layers are indicated in Table.1 in the Results section. While for CIFAR100 and sequence of heterogeneous datasets, experiments are conducted using two convolutional layers with size 16 and 32, followed by two fully connected layers with hidden size indicated in Table.2, 3 and 4. Coreset size are chosen based on the coreset size of other baseline models in MNIST experiments. 200, 40 and 20 samples per task is used as coreset in Permuted MNIST, Multihead Split MNIST and Task Agnostic (Singlehead) Split MNIST experiment respectively. Coreset size of 256 is used in CIFAR100 demonstrations and 512 is used for sequence of heterogeneous datasets.

It must be noted that the two hyperparameters of \VIPARR, namely, the $\beta_k$ and $\gamma_k$ are initialized with the same value for all $k=1,\ldots,o$ layers in this study. Hence the $\beta_k$ and $\gamma_k$ can be replaced with $\beta$ and $\gamma$. In all experiments, $0.08<\gamma<0.3$, as observed from the study in Section 5.1.2, and  $1e^{-09}<\beta<{1e-02}$. The actual choice for each experiment is shown in Table \ref{tab:hyper}.

\begin{table*}[htb]
\begin{center}
\renewcommand{\arraystretch}{1.1}
\caption{Hypermarameters: $\beta$ and $\gamma$ } \label{tab:hyper}
\begin{tabular}{lcccccc}
\hline
\hline
\textbf{}     & \textbf{pMNIST}  & \textbf{MH-}  & \textbf{TA-} & \textbf{MH-}  & \textbf{TA-} &\textbf{Sequence}\\ 
\textbf{} &\textbf{} &\textbf{sMNIST} &\textbf{sMNIST}  &\textbf{CIFAR100} &\textbf{CIFAR100} &\textbf{}
\\ \hline
$\beta$     & $e^{-5}$     & $e^{-5}$    &$e^{-2.5}$   &$e^{-4}$ &$e^{-2}$  &$e^{-9}$\\ \hline
$\gamma$    & $0.12/0.3$   &$0.2$  &$0.1$ &$0.15$ &$0.15$ &$0.08$\\ \hline
\end{tabular}
\end{center}
\end{table*}